# Abstracting Geo-specific Terrains to Scale Up Reinforcement Learning

**Volkan Ustun, Soham Hans, Rajay Kumar**
**USC Institute for Creative Technologies**
**Playa Vista, CA**
**{ustun, sohamhan, kumar}@ict.usc.edu**

**Yunzhe Wang**
**USC Department of Computer Science**
**Los Angeles, CA**
**yunzhewa@usc.edu**

## ABSTRACT

Multi-agent reinforcement learning (MARL) is increasingly ubiquitous in training dynamic and adaptive synthetic characters for interactive simulations on geo-specific terrains. Frameworks such as Unity's ML-Agents help to make such reinforcement learning experiments more accessible to the simulation community. Military training simulations also benefit from advances in MARL, but they have immense computational requirements due to their complex, continuous, stochastic, partially observable, non-stationary, and doctrine-based nature. Furthermore, these simulations require geo-specific terrains, further exacerbating the computational resources problem. In our research, we leverage Unity's waypoints to automatically generate multi-layered representation abstractions of the geo-specific terrains to scale up reinforcement learning while still allowing the transfer of learned policies between different representations. Our early exploratory results on a novel MARL scenario, where each side has differing objectives, indicate that waypoint-based navigation enables faster and more efficient learning while producing trajectories similar to those taken by expert human players in CSGO gaming environments. This research points out the potential of waypoint-based navigation for reducing the computational costs of developing and training MARL models for military training simulations, where geo-specific terrains and differing objectives are crucial.

## ABOUT THE AUTHORS

**Volkan Ustun** is the Associate Director of the Human-Inspired Adaptive Teaming Systems Group at the USC Institute for Creative Technologies. His research augments Multi-agent Reinforcement Learning (MARL) models, drawing inspiration from operations research, human judgment and decision-making, game theory, graph theory, and cognitive architectures to better address the challenges of developing behavior models for synthetic characters, mainly in military training simulations.

**Soham Hans** is a researcher at the USC Institute for Creative Technologies. His research interests are Reinforcement Learning and Large Language Models.

**Rajay Kumar** is a research programmer at the USC Institute for Creative Technologies. He integrates artificial intelligence models with realistic environments.

**Yunzhe Wang** is a Computer Science Ph.D. student at the University of Southern California. His research leverages generative AI to augment human decision-making.

# Abstracting Geo-specific Terrains to Scale Up Reinforcement Learning

**Volkan Ustun, Soham Hans, Rajay Kumar**
**USC Institute for Creative Technologies**
**Playa Vista, CA**
**{ustun, sohamhan, kumar}@ict.usc.edu**

**Yunzhe Wang**
**USC Department of Computer Science**
**Los Angeles, CA**
**yunzhewa@usc.edu**

## INTRODUCTION

The development of autonomous synthetic characters not only presents opportunities but also holds the potential to improve virtual training in interactive simulations radically. These characters can provide safe, replicable, and cost-efficient training, making the future of virtual training simulations promising. For effective training, such simulations require dynamic and adaptive synthetic agents as teammates or respectable opponents that can achieve human-level performance and strategy. However, synthetic entities in many current military simulations have limited behaviors. They usually employ rule-based/reactive computational models with minimal intelligence and cannot adapt based on experience. Experiential learning is essential for developing dynamic and adaptive synthetic characters with credible behavior because attempting to achieve realistic and convincing behavior solely through strict behavioral control would be highly impractical. A strict behavior control approach would require creating rules for each unique task, environment, and potential interaction, making the task nearly impossible. Observation-based behavior model adaptation leverages machine learning utilizing the experience of synthetic entities in combination with appropriate prior knowledge to address the issues in the existing computational behavior models. For example, Reinforcement Learning (RL) can produce adaptive behavior in virtual environments, utilizing the interactions of the synthetic characters with game-like environments in virtual simulations to search for optimal policies.

Multi-agent Reinforcement Learning (MARL) is a promising methodology for developing dynamic and adaptive synthetic characters for interactive simulations. It models multiple agents that learn by dynamically interacting with an environment and each other, providing a framework for evaluating competitive and collaborative dynamics between these agents (Buşoniu et al., 2010). MARL is increasingly ubiquitous in training dynamic and adaptive synthetic characters for interactive simulations (Zhang et al., 2021). MARL algorithms have shown the ability to learn policies demonstrating cooperative and strategic team-oriented behaviors (Gronauer & Diepold, 2022). Frameworks such as Unity's ML-Agents (Juliani et al., 2018) help to make MARL experiments more accessible to the simulation community, and virtual simulation environments like RIDE (Hartholt et al., 2021) assist in running these simulations on geo-specific terrains.

Advances in MARL have also proven beneficial for military training simulations, but they come with immense computational requirements due to their complex, continuous, stochastic, partially observable, non-stationary, and doctrine-based nature. Furthermore, these simulations require geo-specific terrains, further exacerbating the computational resources problem. However, waypoint-based navigation has shown promise in addressing these computational requirements (Aris et al., 2023). By replacing the fine-grained action space with a more abstracted, navmesh-based waypoint movement system in single-agent reinforcement learning, such a movement system can discretize the movement of agents and potentially increase the generality and success rate of the models. For example, (Koresh et al., 2024) demonstrated efficiency in learning and improved performance with waypoint-based navigation on realistic, but not geo-specific, terrains.

This paper augments the waypoint-based navigation to multi-agent systems on geo-specific terrains. It confirms that leveraging waypoints can decrease the computational requirements of MARL training with superior performance compared to fine-grained action spaces. While demonstrating the efficacy of waypoint-based navigation, we utilize a novel MARL scenario, where each side has a different objective, taking place on geo-specific terrain. Differing objectives are not a common scenario type in classical MARL experiments where each side has a similar goal. However, differing objectives are an essential feature for many training environments, where each side has a different perspective; for example, one could defend an area, whereas the other side tries to establish a position in the defended area. We show that waypoint-based navigation supports faster and more efficient learning in such scenarios.
Secondly, we compare waypoint-based navigation with real human trajectories playing the Counter-Strike: Global Offensive (CSGO) game. Our comparisons and analysis show that waypoint-based navigation allows trajectories very similar to those taken by expert human players, providing further evidence of the validity of waypoint-based representation for modeling navigation on geo-specific terrains.

After providing a short background, we introduce our waypoint-based movement system. Next, we analyze the use of waypoints for game-play human trajectories. We then introduce differing objective scenarios on a geo-specific terrain and present our results on the efficacy of waypoint-based navigation in MARL experiments. We conclude with the potential implications of waypoint-based movement systems for military training simulations.

## BACKGROUND

Deep MARL models blend advances in Artificial Neural Networks with algorithms from MARL research, yielding excellent results for toy problems like predator-prey, cooperative navigation, and physical deception (Lowe et al. 2017). More recently, applications of deep MARL models have shown great promise in more complex environments, such as soccer simulations (Kurach et al. 2019) and first-person shooters (Jaderberg et al. 2019), devising novel and effective behavior policies. Many state-of-the-art MARL algorithms utilize an actor-critic approach (Konda & Tsitsiklis, 1999). In this approach, during training time, a central critic can observe all the actors (agents) and their rewards. As a result, it can inform individual agent policies, potentially yielding a learned consensus in cooperative tasks (Lowe et al., 2017). For example, modifications to the popular on-policy single agent Proximal Policy Optimization (PPO) (Schulman et al., 2017) algorithm for multi-agent settings under actor-critic paradigm can be surprisingly effective (Yu et al., 2022). Multi-agent Deep Deterministic Policy Gradient (MADDPG) (lowe2017multi), which is a multi-agent algorithm by design, delivers excellent results for toy problems like predator-prey, cooperative navigation, and physical deception. (Ustun et al., 2022) utilizes such MARL techniques to concurrently learn policies for opposing forces for military training scenarios at the squad level.

Counterfactual multi-agent policy gradient (COMA) (Foerster et al., 2018) is another actor-critic architecture that tackles the challenge of multi-agent credit assignment in cooperative settings with a unique shared reward through counterfactuals. Multi-agent Posthumous Credit Assignment (MA-POCA) (Cohen et al., 2021) extends COMA via attention to better handle the credit assignment with terminated agents in training episodes. Before MA-POCA, the most common solution for eliminated agents was to account for the maximum number of agents and place the inactive agents in an absorbing state, allowing for reward to propagate back to eliminated agents. MA-POCA approaches the posthumous credit assignment problem via attention rather than a fully connected neural network, avoiding the need for any absorbing state while accurately quantifying an individual’s contribution to the team’s outcome. MA-POCA performed better than PPO and COMA in various cooperative environments, and the margin was most significant in scenarios that added or removed agents during a simulation. These results are auspicious for MARL experiments with military training scenarios, where agent elimination is prevalent and self-sacrificing behavior can benefit the team's outcome. (Koresh et al., 2024) discusses MARL experiments with MA-POCA where agents have heterogeneous capabilities.

In earlier work (Koresh et al., 2024), we leveraged waypoint-based navigation graphs for unit maneuvers at different resolutions on realistic terrains, enabling the simulation execution up to 30 times faster than in real-time. (Koresh et al., 2024) also tested the waypoint-based agents against their fine-grained moving counterparts for a more direct comparison in head-to-head matches. As shown in Figure 1, waypoint-based systems performed better than their fine-grained movement counterparts over 100 evaluations in each of the four scenarios. This paper builds on this earlier work and augments it for geo-specific terrains and provides comparisons against human trajectories collected in game play.

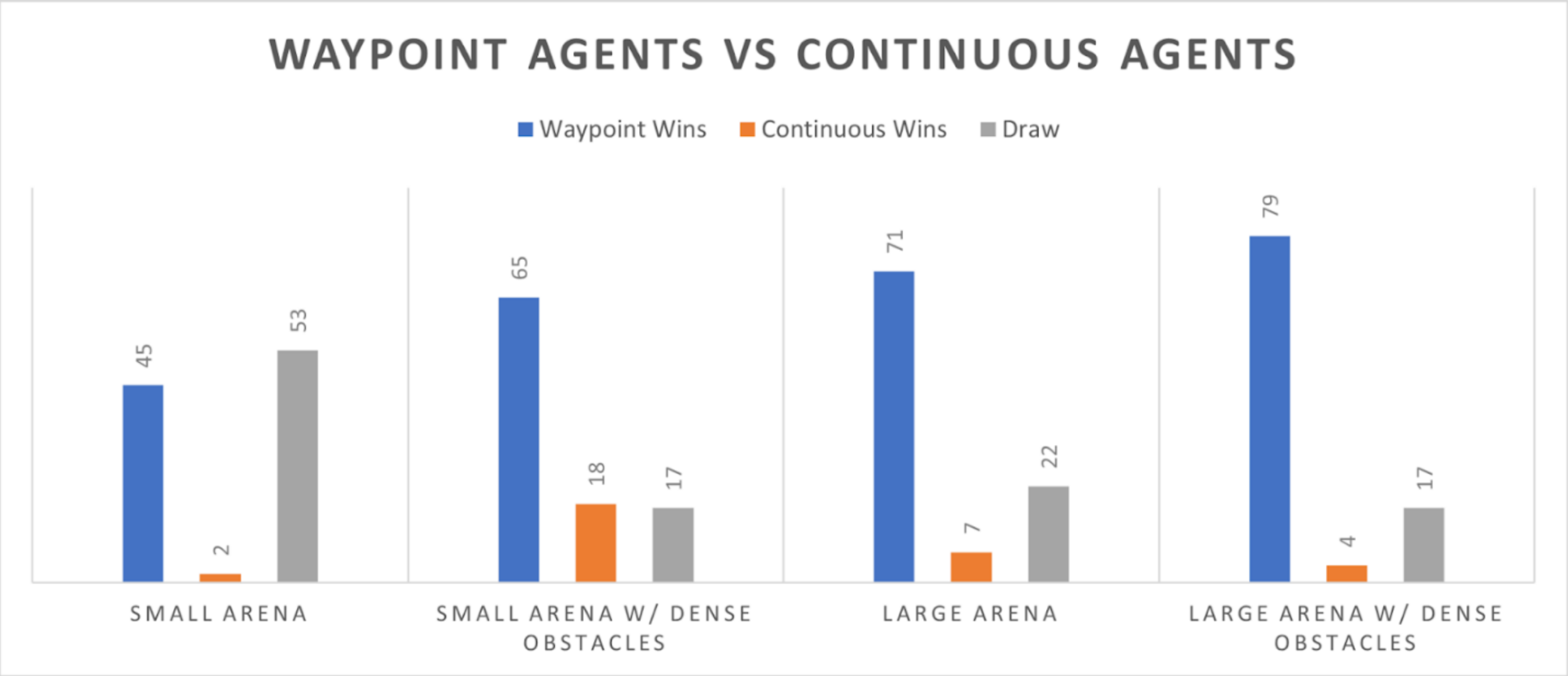

**Figure 1. Waypoint-based vs fine-grained (continuous) wins in head-to-head matches**

## WAYPOINTS

We developed an automatic waypoint generation system to create a waypoint-based movement graph for a terrain used in a scenario. We can generate waypoints for both geo-specific and basic Unity terrains with this system; for instance, the examples in this paper are for geo-specific Unity terrains and indoor environments. Before deploying the waypoints, we create a NavMesh for the terrain with constraints such that the agent cannot ascend steep slopes. The action space used in our experiments involves choosing which of the eight adjacent waypoints (cardinal directions and diagonals) as the move destination. We laid out a grid of waypoints one at a time, moving from the southwest corner to the northeast corner and assigning connections to any waypoints in the southeast, south, southwest, and east directions.

When a waypoint is visited, the environment attempts to create adjacent waypoints in any direction that does not have one at a parameterized distance. However, not all generated waypoints and edges are valid. A waypoint is marked invalid if there is no proper NavMesh position at that location. An edge is marked invalid if there is no good path on the NavMesh to get to that waypoint, the waypoints are too far apart vertically, or the path is far longer than the Euclidean distance between the waypoints. While this approach covers most of the terrain, it does leave some key areas bare, especially for indoor settings, such as underpasses. We utilize a breadth-first search to automatically fill the waypoint graph gaps. Furthermore, we developed an additional capability to fine-tune waypoints and their connections, utilizing input from human experts when available.

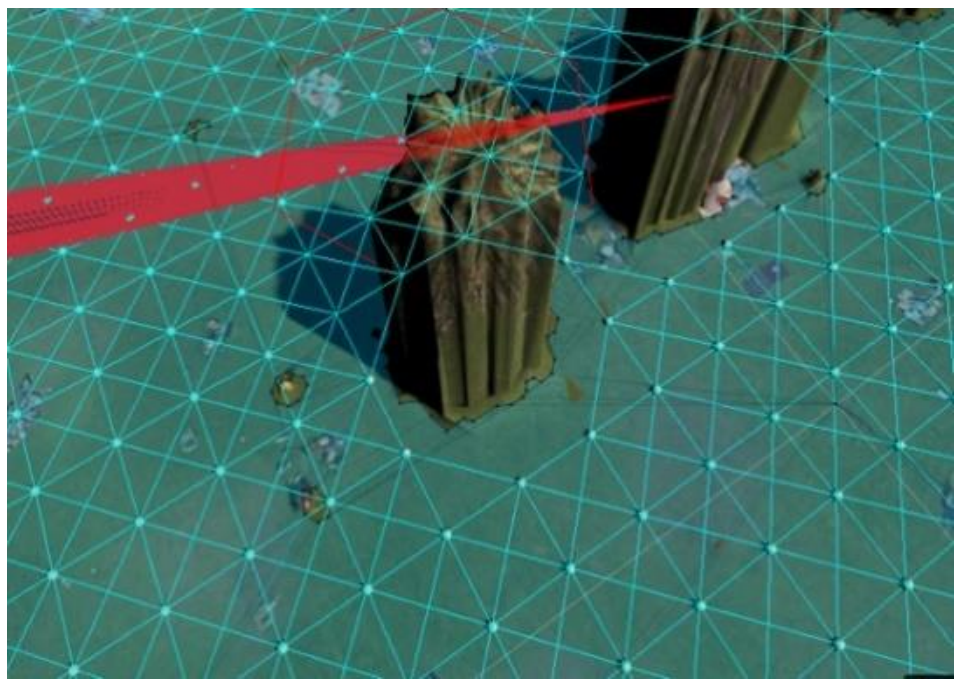

**Figure 2. Waypoints**

Our RL experiments will utilize the generated waypoint-based movement graph. If an edge or waypoint is marked invalid, the action masking will prune the associated action, ensuring the system effectively handles invalid edges and waypoints. Our experiments treat all edges as equidistant, although the diagonal edges are slightly longer. Figure 2 depicts the waypoint placement.

This paper compares the performance of waypoint and non-waypoint agents. The main difference between them is their action spaces. Non-waypoint agents use fine-grained actions along with a branch to adjust their rotation as defined in the original Unity dodgeball scenario (Berges at al., 2021). However, we removed the dash mechanic, and dashes are not part of the action spaces for either agent type. For movements, the non-waypoint-based agents use the original implementation, which features two fine-grained branches representing movement along the X and Z directions.

On the other hand, our waypoint-based agents use one discrete action branch consisting of 9 possible actions. Eight of these actions correspond to the eight cardinal directions, and the ninth is for standing still. As stated earlier, movement actions are masked between waypoints when one or more directions are inaccessible, reducing the search space without affecting gameplay.

## HUMAN TRAJECTORIES VS WAYPOINT REPRESENTATIONS

Our waypoint representation effectively captures human trajectories within a game-specific geo-terrain environment. Utilizing the game Counter-Strike: Global Offensive (CSGO) as our test bed, we leverage its publicly accessible tournament demo file to obtain human trajectory data. We then quantitatively demonstrate that our waypoint-based navigation model can accurately replay the fine-grained human player trajectories in the actual game with minimal deviation. Such replays confirm the precision and reliability of our waypoint system in modeling real-world human paths within a digital game context.

To obtain the human trajectory data, we employed the Esports Trajectories & Actions (ESTA) dataset (Xenopoulos et al., 2022), which includes an extensive collection of parsed CSGO demos from tournament matches between January 2021 and May 2022. Each CSGO match consists of two teams of five players, with up to 30 rounds where players navigate a fixed map to complete objectives within 155 seconds. We focused on the Dust2 map and processed the ESTA dataset to extract player movement sequences for individual rounds across all matches, with 197 matches, 5178 rounds, and 51780 human trajectory data instances. Each trajectory instance is a time series capturing a player's path over one round in x, y, and z coordinates, with a sampling frequency of 0.5 seconds per data point. An example trajectory is shown in Figure 3 (left).

Before aligning human trajectories with our waypoint representation, we first recreated the Dust2 map environment in Unity by extracting the map file from the real downloaded game and decompiling it into mesh objects using the open-source tool BSPSource (BSP Source, 2024); this ensures the game objects, such as roads and obstacles, in our recreated version has the exact placement and scale as in the original game. We then imported the decompiled map in mesh objects into Unity and generated waypoints. To align human trajectories with the generated waypoints, we identified the closest waypoint for each player position data point in Unity using Euclidean distance shown in Figure 3 (middle). Given that our waypoint representation has a finer spatial resolution than the recorded human data, some waypoints were skipped between sample points. To address this, we employed Dijkstra's shortest path algorithm to fill in the skipped waypoints, creating a fully connected path in the waypoint representation that mirrors actual human trajectories as depicted in Figure 3 (right).

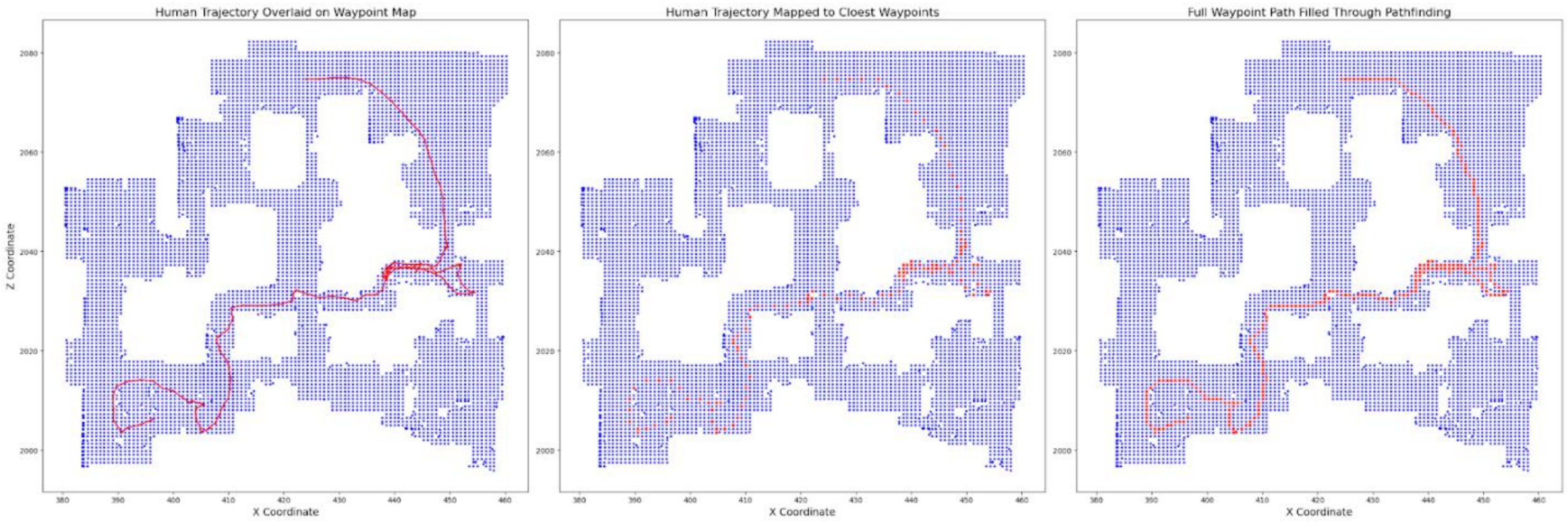

**Figure 3: An example human trajectory in the Dust2 map of CSGO, from T-spawn to A-Bombsite.**
Blue dots represent all waypoints; red lines and dots show the path. Panels: (left) the original human trajectory, (middle) trajectory mapped to closest waypoints, and (right) complete waypoint path via pathfinding.

Lastly, we quantitatively evaluated the accuracy of our waypoint representation by measuring the deviation between the distance traveled in our replayed waypoint and the actual human trajectories. We used two metrics for comparison: stepwise difference (0.5-second intervals) and round-wise difference (total distance traveled in one round). Our findings show strong alignment, with R-squared values of 0.76 for stepwise and 0.96 for round-wise differences (Figure 4). Additionally, we calculated the relative difference as:

$$\boldsymbol{Relative\ Difference} = \frac{|\boldsymbol{Euclidean\ Distance} - \boldsymbol{Waypoint\ Distance}|}{\textbf{Euclidean Distance}}$$

Our relative difference distribution analysis showed that, on average, our waypoint representation deviates by 6.8% from the actual trajectories, with a standard deviation of 0.04.

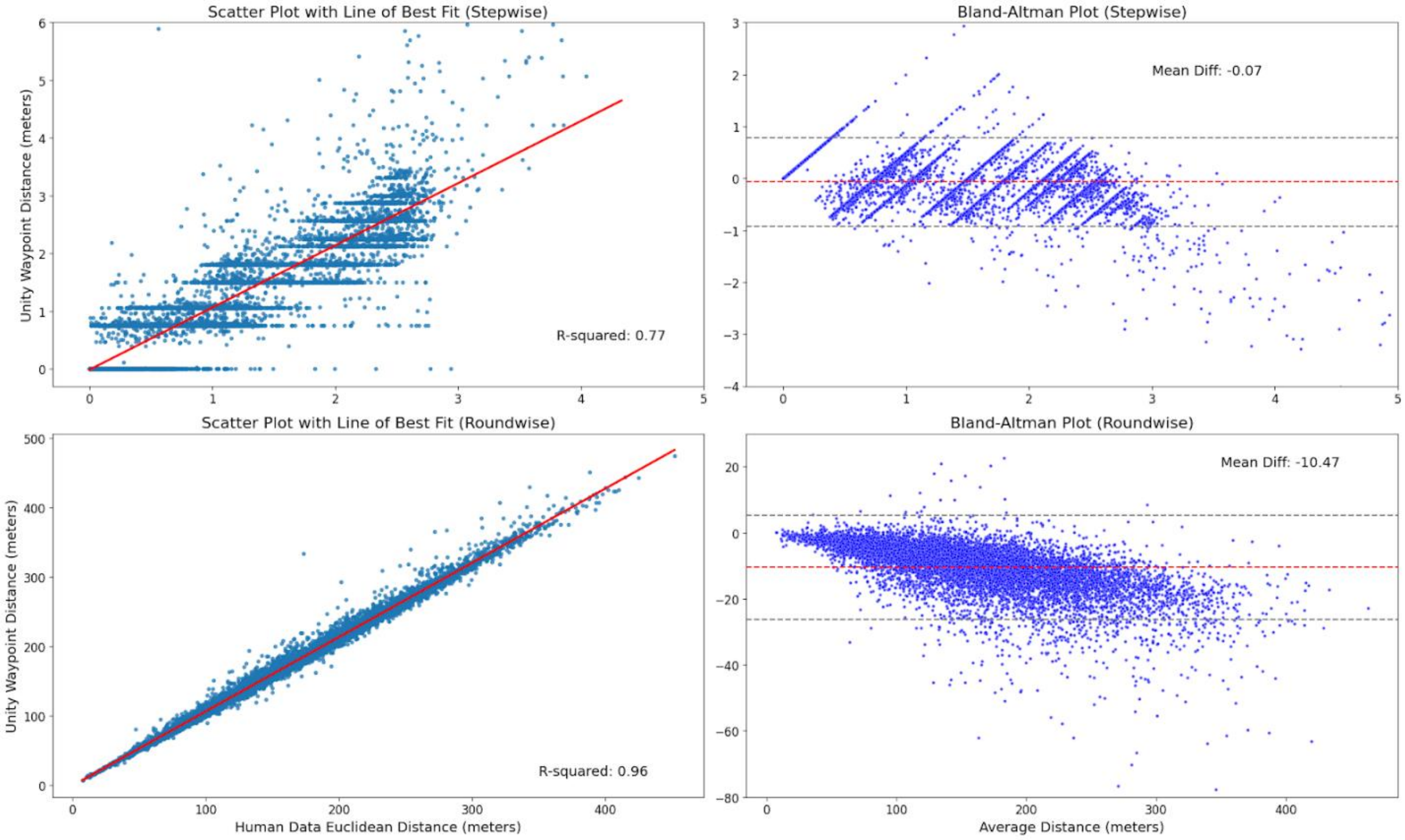

**Figure 4: Accuracy of Waypoint Representation.**

Scatter and Bland-Altman plots demonstrate the accuracy of our waypoint representation in capturing human trajectories. (top) Stepwise distances show a strong correlation (R-squared = 0.77) and a mean difference of -0.07. (bottom) Round wise distances exhibit an excellent correlation (R-squared = 0.96) and a mean difference of -10.47. These results confirm the precision and reliability of our waypoint system in modeling real-world human paths with minimal deviation.

## SIMPLE COMBAT SCENARIOS ON GEO-SPECIFIC TERRAINS WITH DIFFERING OBJECTIVES

This section describes the proof-of-concept military scenario on geo-specific terrain, which is used to test the performance of waypoint and non-waypoint agents. The terrain for these experiments is a section of the Razish village area at the National Training Center. The predominantly flat boundary of this region measures 91.30 x 41.25 units and is strategically designed to challenge the agents. It includes 1812 waypoints, facilitating the movement of waypoint-based agents.

The combat scenario involves two teams, the Red and Blue teams, each comprising four agents. The Blue Target destination, a pivotal point, is strategically located near the Red team's starting locations, adding to the intensity of the scenario (Figure 5).

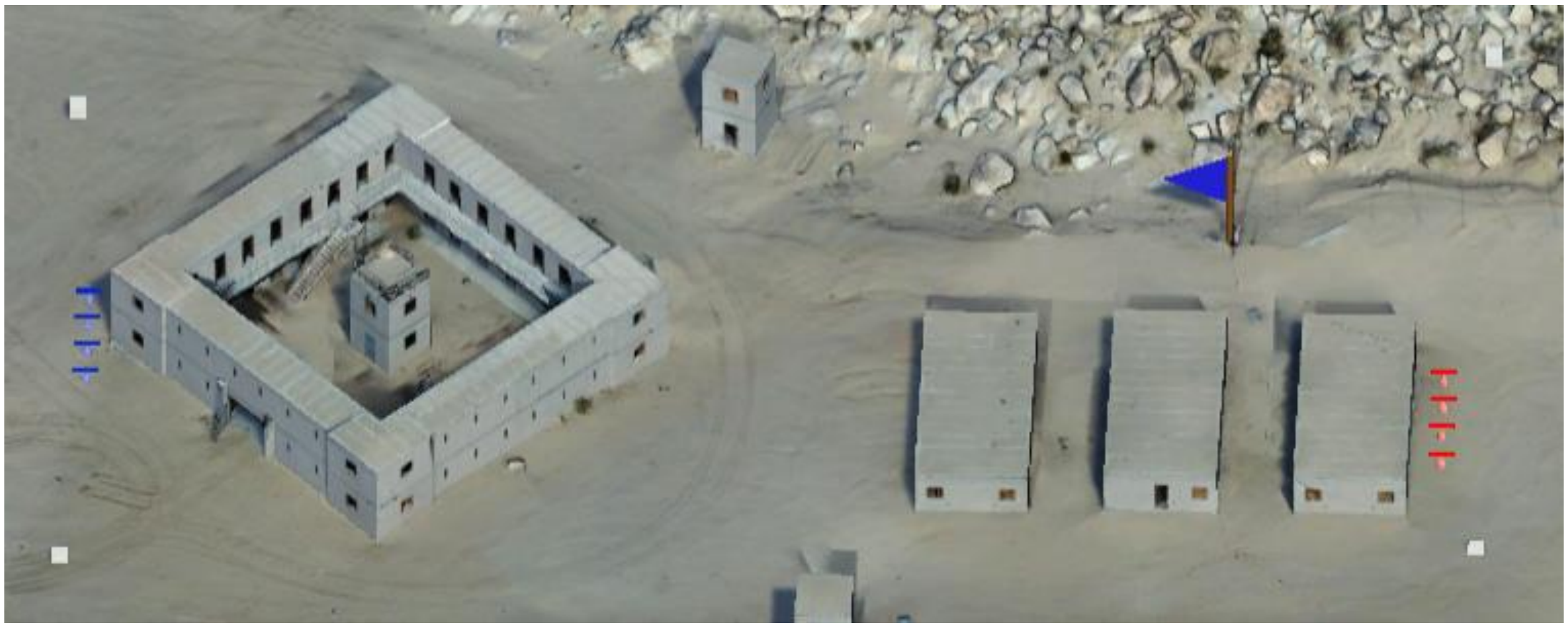

**Figure 5: A simple combat scenario location at the Razish Village, National Training Center**

The agents' objective in the Blue team is to try to reach the Blue Target destination. If any one of the agents can make it to the Blue Target, then the Blue Team wins. The purpose of the Red Team is to defend the Blue Target area and eliminate any Blue Team agent who tries to reach the target.

The agents in each team can shoot at each other, and getting shot more than five times eliminates the agent and removes it from the scenario. We provide an aim assistance mechanism to the scenario agents to assist with shooting accuracy, allowing them to target enemies automatically. Some noise in the form of jitter is added into this mechanism to ensure that the shot's accuracy is affected by the distance to the target.

**Experiment Design and Results**

For our experiments, we have chosen ten starting positions for the Red team (Red Starts) and ten other starting positions for the Blue team (Blue Starts). All of the Red Start positions lie on the right half of the map shown in Figure 5, and all of the Blue Starts lie on the left half of the map.

We used the MA-POCA algorithm (Cohen et al., 2021) with self-play in 2 setups for our training experiments.

- First, we trained a scenario in which both the Red and Blue teams used waypoint-based movement, and their initial positions were chosen randomly from the set of starting positions.
- The second setup used agents with fine-grained (continuous) movements on both teams with randomly chosen starting positions.

We use the ELO score metric for each team to assess training performance. The ELO score represents the relative skill of the agents compared to previous versions of themselves, which is popularized by (Silver et al., 2017) as a metric for evaluation of progress in reinforcement learning experiments. Figure 6 depicts ELO score progress during training for waypoint movements and fine-grained (continuous or non-waypoint) movements. As seen in Figure 6, waypoint-based movements reach significantly higher ELO scores.

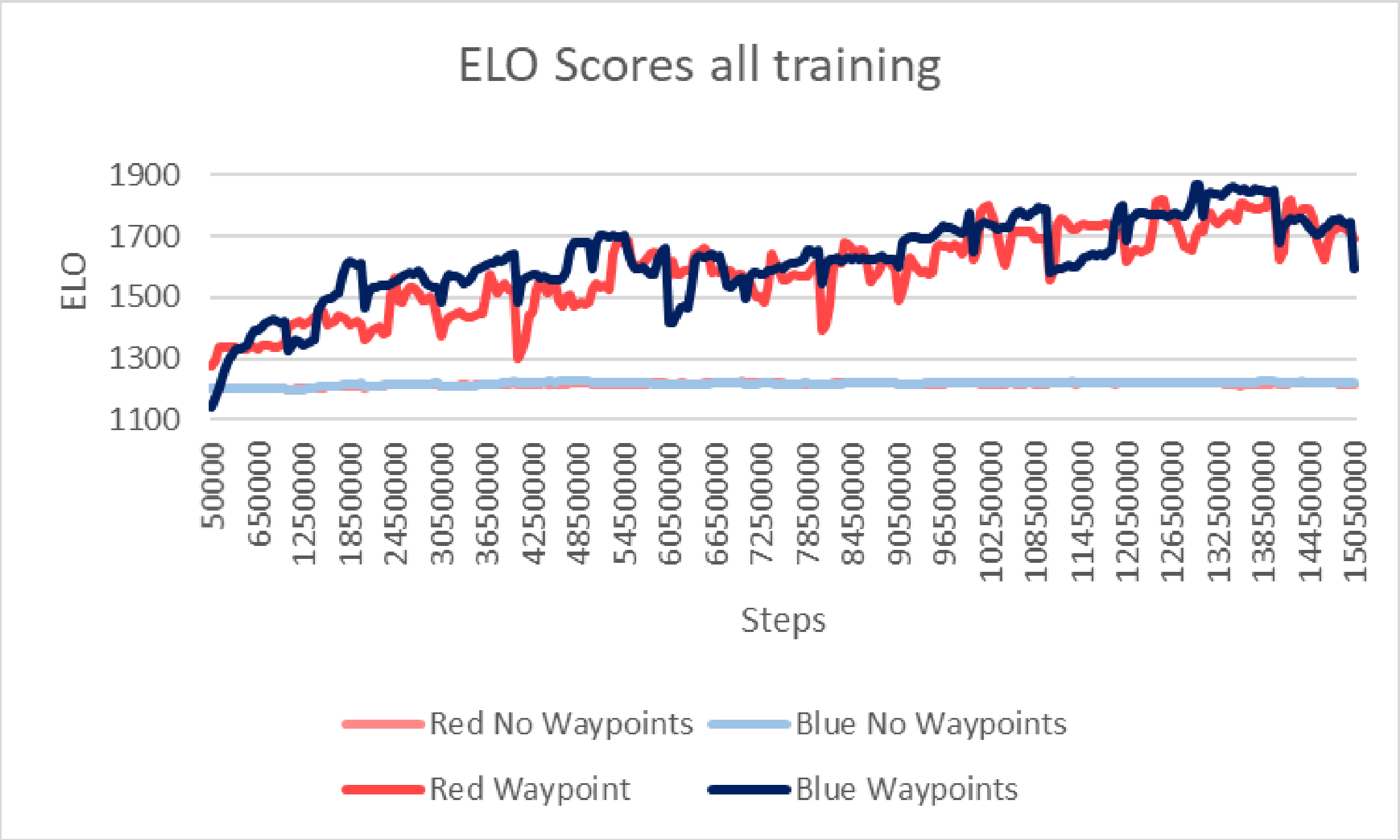

**Figure 6: ELO Scores for waypoint and fine-grained (non-waypoint) movements**

Our training experiments resulted in the creation of four distinct trained agent teams: (1) Blue Team with Waypoints, (2) Red Team with Waypoints, (3) Blue Team with Fine-grained movements, and (4) Red Team with Fine-grained movements. These teams were then pitted against each other in a series of head-to-head matches, forming a comprehensive evaluation of the impact of waypoint-based movements. Each combat experiment involved 100 combinations of initial positions for ten starting positions for each team.

When the Blue team with Waypoints is against the Red team with Waypoints, the Blue team wins 70 matches out of 100, showing that the Blue Team has a slight advantage over the Red Team. When both teams use fine-grained movements, the Blue team still wins 55 matches out of 100. This partiality is due to the imbalance in the game objectives design, which favors the Blue Team slightly. The most likely explanation for this imbalance is that the Blue Team usually has access to a better cover site in the diamond-shaped building and can get into the cover position faster. Once covered, the Blue Team can eliminate the entire Red Team more effectively before traveling to the Blue Target destination. Still, the Red Team can win 30 out of 100 rounds when they utilize the waypoint-based movements.

However, when the Red team with Fine-grained movements was pitted against the Blue team with waypoints, the Blue team completely dominated the matchup, winning 98 out of 100 matches. Further analysis of these matches showed us that the waypoints-based movements allowed the Blue Team to coordinate their movements much more efficiently while maintaining effective formations and eliminating the Red Team one by one.

On the other hand, when the Red Team with waypoints-based movements faced the Blue Team with fine-grained movements, the Red Team won 88 matches out of 100. Even though the scenario favors the Blue Team, waypoint-based movement capabilities allowed the Red Team to significantly outperform their opponents with fine-grained movements. Waypoint-based movements endowed the Red Team with more effective coordination of movements, allowing them to efficiently target and eliminate opponents one at a time. These results are comparable to the ones reported by (Koresh et al., 2024), and they further validate the effectiveness of waypoint-based movements.

## DISCUSSION

A waypoint-based movement system offers several advantages. Firstly, it can accelerate learning in MARL experiments, enabling faster training of behavior policies. Secondly, learned policies exhibit robustness when transferred back to the original action space. Thirdly, automatic waypoint generation accommodates geo-specific terrains, facilitating MARL experiments on realistic terrains. Additionally, it establishes a foundation for making further assumptions based on a project's unique design and needs. For instance, waypoint-based movement graphs can serve as a basis for utilizing simpler graph-based environments, providing opportunities for even faster training and subsequent transfer back to higher fidelity environments for fine-tuning.

Training units with differing objectives is crucial in military simulations. Waypoint-based movement systems facilitate the convergence of MARL experiments with differing objectives. In a proof-of-concept scenario, we learned effective policies simultaneously for units with differing objectives, such as accessing an area and defending that area.

While our results are promising, explorations with further abstractions can lead to even faster experiments. Such abstractions depend mainly on the nuances of each project. Our focus was primarily on accelerating experiments in realistic environments, particularly those with geo-specific terrains. However, our findings suggest that abstractions could significantly reduce computational requirements for similar experiments, with waypoints providing a solid foundation for creating high-speed abstract simulation environments.

In future work, we aim to incorporate additional functionality and units, such as vehicles or medical units, to create a broader range of military scenarios. More importantly, we will continue our efforts to understand the effectiveness of abstracted environments for performance and their connections to waypoint-based representations. Such abstractions offer an avenue for faster reinforcement learning, which is crucial for complex systems like military training simulations. Reasonable abstractions with pathways for transfer learning should help narrow the search space without compromising critical strategic components like cooperative positioning, ultimately enabling the generation of intelligent behavior in high-fidelity environments.

## CONCLUSION

Waypoint-based movement systems speed up learning in MARL for scenarios taking place on geo-specific terrains and assist in generating transferable policies while allowing trajectories like the ones taken by human players. They also provide a foundation for abstractions that can further accelerate training. While our results are promising, further abstractions can lead to even faster experiments, especially in complex environments like military training simulations. Future work will focus on incorporating additional functionality and units to create a broader range of scenarios and understanding the effectiveness of abstracted environments for performance.

## ACKNOWLEDGMENTS

Research was sponsored by the Army Research Office and was accomplished under Cooperative Agreement Number W911NF-14-D-0005. The views and conclusions contained in this document are those of the authors and should not be interpreted as representing the official policies, either expressed or implied, of the Army Research Office or the U.S. Government. The U.S. Government is authorized to reproduce and distribute reprints for Government purposes notwithstanding any copyright notation.